\documentclass[11pt,a4paper]{article}
\usepackage{acl2015}
\usepackage{times}
\usepackage{url}
\usepackage{latexsym}
\usepackage[english]{babel}
\usepackage{amsmath}

\usepackage{graphicx} 
\graphicspath{{figures/}} 

\usepackage{booktabs}
\usepackage{multirow}

\title{FAQ-based Question Answering via Word Alignment}
\author{Zhiguo Wang \\
  IBM T.J. Watson Research Center \\
  Yorktown Heights, NY, USA \\
  {\tt zhigwang@us.ibm.com} \\ %
  \And
  Abraham Ittycheriah \\
  IBM T.J. Watson Research Center \\
  Yorktown Heights, NY, USA \\
  {\tt abei@us.ibm.com} \\}
\date{}

\begin{document}
\maketitle
\begin{abstract}
In this paper, we propose a novel word-alignment-based method to solve the FAQ-based question answering task. First, we employ a neural network model to calculate question similarity, where the word alignment between two questions is used for extracting features. Second, we design a bootstrap-based feature extraction method to extract a small set of effective lexical features. Third, we propose a learning-to-rank algorithm to train parameters more suitable for the ranking tasks. Experimental results, conducted on three languages (English, Spanish and Japanese), demonstrate that the question similarity model is more effective than baseline systems, the sparse features bring 5\% improvements on top-1 accuracy, and the learning-to-rank algorithm works significantly better than the traditional method. We further evaluate our method on the answer sentence selection task. Our method outperforms all the previous systems on the standard TREC data set.
\end{abstract}

\section{Introduction}
Question Answering (QA) aims to automatically understand natural language questions and to respond with actual answers. The state-of-the-art QA systems usually work relatively well for factoid, list and definition questions, but they might not necessarily work well for real world questions, where more comprehensive answers are required. Frequently Asked Questions (FAQ) based QA is an economical and practical solution for general QA \cite{burke1997question}. Instead of answering questions from scratch, FAQ-based QA tries to search the FAQ archives and check if a similar question was previously asked. If a similar question is found, the corresponding answer is returned to the user. The FAQ archives are usually created by experts, so the returned answers are usually of higher-quality.

The core of FAQ-based QA is to calculate semantic similarities between questions. This is a very challenging task, because two questions, which share the same meaning, may be quite different at the word or syntactic level. For example, ``How do I add a vehicle to this policy?" and ``What should I do to extend this policy for my new car?" have few words in common, but they share the same answer. In the past two decades, many efforts have been made to tackle this lexical gap problem. One type of methods tried to bridge the lexical gap by utilizing semantic lexicons, like WordNet \cite{burke1997question,wu2005domain,yang2007ontology}. Another method treated this task as a statistical machine translation problem, and employed a parallel question set to learn word-to-word or phrase-to-phrase translation probabilities \cite{berger1999information,jeon2005finding,lee2008bridging,xue2008retrieval,bernhard2009combining,zhou2011phrase}. Both of these methods have drawbacks. The first method is hard to adapt to many other languages, because the semantic lexicon is unavailable. For the second method, a large parallel question set is required to learn the translation probabilities, which is usually hard or expensive to acquire. To overcome these drawbacks, we utilize distributed word representations to calculate the similarity between words, which can be easily trained by only using amount of monolingual data.

In this paper, we propose a novel word-alignment-based method to solve the FAQ-based QA tasks. The characteristics of our method include: (1) A neural network model for calculating question similarity with word alignment features. For an input question and a candidate question, the similarities of each word pairs (between the two questions) are calculated first, and then the best word alignment for the two questions is computed. We extract a vector of dense features from the word alignment, then import the feature vector into a neural network and calculate the question similarity in the network's output layer. (2) A bootstrap-based feature extraction method. The FAQ archives usually contain less than a few hundred questions, and in order to avoid overfitting, we are unable to use too many sparse features. Therefore, we come up with this method to extract a small set of effective sparse features according to our system's ranking results. (3) A learning-to-rank algorithm for training. The FAQ-based QA task is essentially a ranking task, our model not only needs to calculate a proper similarity for each question pair, but also needs to rank the most relevant one on top of the other candidates. So we propose a learning-to-rank method to train parameters more suitable for ranking. Experimental results, conducted on FAQ archives from three languages, demonstrate that our method is very effective. We also evaluate our method on the answer sentence selection task. Experimental results on the standard TREC data set show that our method outperforms all previous state-of-the-art systems.

\begin{figure}[tbp]
\begin{center}
\includegraphics[width=0.5\textwidth]{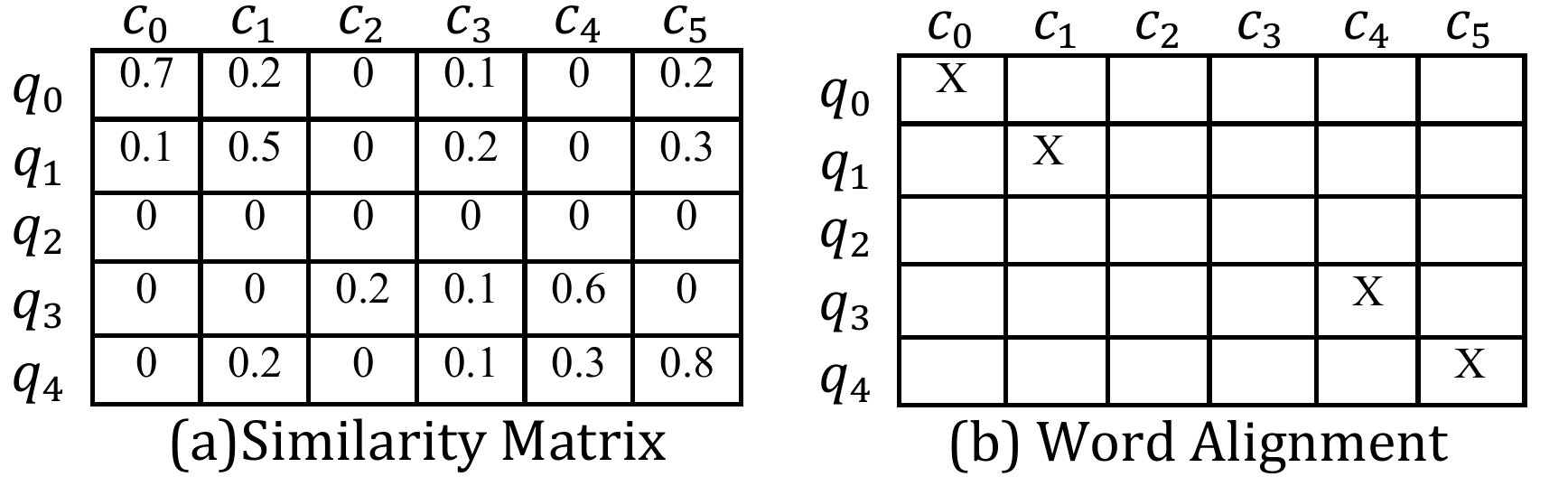}
\end{center}
\caption{Word alignment example.}
\label{fig:wordalign}
\end{figure}

\section{Method}
The task of FAQ-based QA is that given a \emph{query} question, rank all the \emph{candidate} questions according to the similarities between two questions. We define the similarity between a query $Q$ and a candidate $C$ as $sim(Q, C) = f(X)$, where $X$ is a feature vector extracted from ($Q$, $C$) pair, and $f(*)$ is a linear or non-linear function. In this work, we represent $f(*)$ as a neural network model, where the input is the feature vector $X$, and the output layer contains only one neuron which predicts the similarity of the two questions. We choose the \emph{sigmoid} activation function for the output layer, so that the similarity is constrained in the range [0, 1]. In order to execute this model, we still have two remaining questions: (1) How to represent the feature vector $X$? (2) How to train the model for ranking?

\subsection{Feature Definition}
For the first question, we propose to extract features from the word alignment between two questions. Let's denote the query question as $Q=q_0,q_1,...,q_i,...,q_m$ and the candidate question as $C=c_0,c_1,...,c_j,...,c_n$, where $q_i$ and $c_j$ represent words in questions. First, we calculate the word similarity between $q_i$ and $c_j$ according to the cosine distance of their distributed representations $v_{qi}$ and $v_{cj}$ 
\begin{align*}
 sim( q_i, c_j)=max(0,cosine(v_{qi},v_{cj}))
\end{align*}
Then, we obtain the similarity matrix for the question pair by calculating similarities of all word pairs, e.g. Figure \ref{fig:wordalign}(a). Finally, we compute the best word alignment for the question pair based on the similarity matrix, e.g. Figure \ref{fig:wordalign}(b) shows the word alignment computed based on Figure \ref{fig:wordalign}(a). 

We define some dense features based on the word alignment. Let's denote the alignment position for each query word $q_i$ as $align_i$, and the corresponding alignment score as $sim_i$. For example, for word $q_3$ in Figure \ref{fig:wordalign}, $align_3=4$ and $sim_3=0.6$. We denote the unaligned word as $unalign_i$. We also take into account the importance of each word by employing its inverse document frequency (IDF) score, and denote it as $idf_i$. We define the following dense features:

\begin{itemize}
\item \textbf{similarity}: $f_0=\sum_i sim_i * idf_i / \sum_i idf_i $. This feature measures the question similarity based on the aligned words.
\item \textbf{dispersion}: $f_1=\sum_i (|align_i-align_{i-1}-1|)^2 $. This feature prefers the candidate where contiguous query words are aligned to contiguous words in the candidate.
\item \textbf{penalty}: $f_2=\sum_{unalign_i}idf_i / \sum_i idf_i$. This feature penalizes candidates based on the unaligned query words.
\item \textbf{5 important words}: $f_{i_{th}}=sim_{i_{th}}*idf_{i_{th}}$. This feature type contains 5 features. Each feature shows the alignment score of the i-th important words, where we evaluate the importance of a word by its IDF score.
\item \textbf{reverse}: Extract above features by swapping roles of query and candidate questions.
\end{itemize}

We also define some spare lexical features. Considering the fact that our FAQ archives contain only less than a few hundred questions, we cannot extract too manny sparse features, otherwise our model will overfit the training set. We design a bootstrap-based feature extraction method to extract a small set of effective lexical features according to the model's ranking results. The input to our method contains a seed model, a FAQ archive and a set of sparse feature templates. For each query question, the workflow includes:

\begin{itemize}
\item \textbf{Step 1}: Rank all the candidates with the seed model. If the rank-1 candidate is relevant, return without doing anything.
\item \textbf{Step 2}: Find the first relevant candidate $C^+$ from the ranking list, and collect all the irrelevant candidates \{$C^-$\} above $C^+$.
\item \textbf{Step 3}: Collect sparse features $F^+$ from $C^+$, and sparse features $F^-$ from \{$C^-$\}. Then, only keep the sparse features occurred in $F^+$ but not occurred in $F^-$.
\end{itemize}

We use this method to extract a group of sparse features, then add these features to the model and retrain the model. This procedure can iterate many times until getting a stable performance. Our feature templates contain: aligned query words, aligned candidate words and aligned query-candidate word pairs. In our experiments, the performance can converge within 10 iterations, and the final model usually contains less than 1,500 sparse features.

\subsection{Learning to Rank Algorithm}
The traditional method for training the similarity model is to cast the task as a binary classification problem \cite{heilman2010tree,severyn2013automatic,yao2013answer}. All the possible question pairs are collected from the training set, and if the question pair is relevant, assign a label ``+1" , otherwise assign a label ``-1". Then the model is trained to optimize the classification accuracy. However, the FAQ-based QA task is essentially a ranking problem. The similarity model not only needs to calculate a proper similarity for each question pair, but also needs to rank the most relevant candidate on top of the others. Therefore, we propose a novel learning-to-rank algorithm to explicitly optimize the ranking (top-1) accuracy. We define the loss function for each query $Q_i$ and all its irrelevant candidates \{$C_j$\} as: 
\begin{align*}
 \label{equ:loss}
 \begin{split}
 l_i = &\underbrace{\sum_{C_j} max(0, \epsilon + sim(Q_i,C_j)-sim(Q_i,{C_j}^*))}_{term_1}\\
 &-\underbrace{sim(Q_i,{C_j}^*)}_{term_2}
 \end{split}
\end{align*}
where $\epsilon$ is a margin, and ${C_j}^*$ is the first relevant candidate in the ranking list. $term_1$ aims to decrease the similarities for the irrelevant candidates ranked above ${C_j}^*$ or below ${C_j}^*$ but with a margin less than $\epsilon$, and $term_2$ aims to improve the similarity of ${C_j}^*$. We utilize the back propagation algorithm \cite{rumelhart1988learning} to minimize the loss function over the training set, and employ the AdaGrad strategy \cite{duchi2011adaptive}. In our experiments, we set $\epsilon$ as 0.03 and the learning rate as 0.1.

\begin{figure*}[tbp]
\begin{center}
\includegraphics[width=0.9\textwidth]{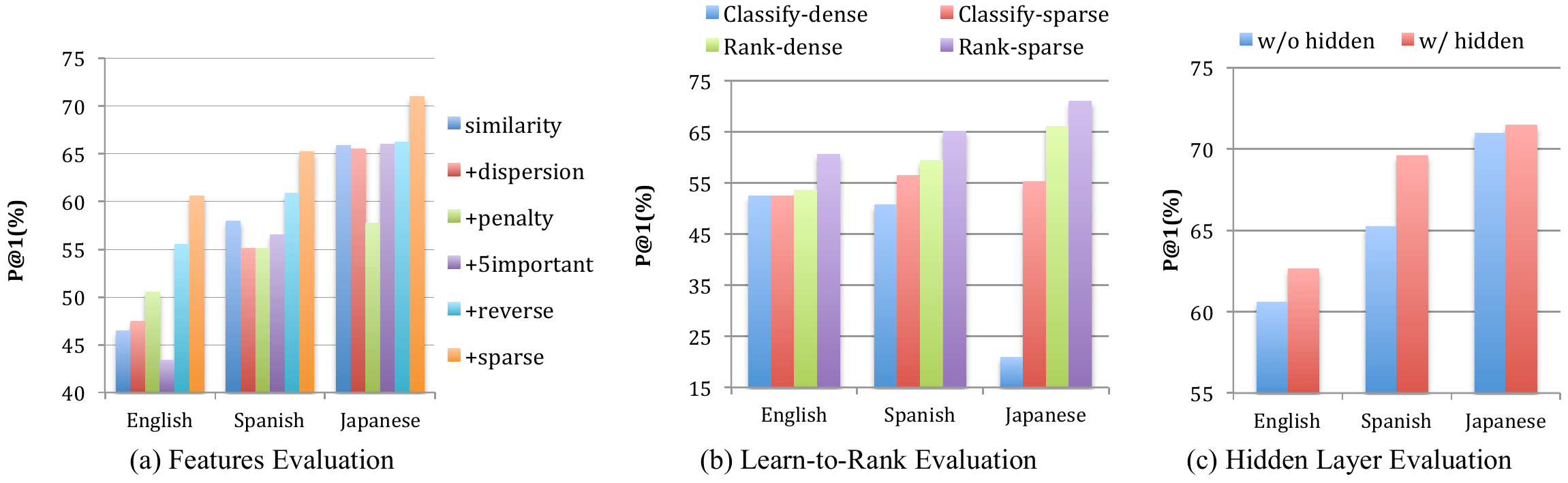}
\end{center}
\caption{Characteristics of our model.}
\label{fig:curves}
\end{figure*}

\section{Experiment}
\subsection{Experimental Setting}
We conducted experiments on FAQ archives from three languages (English, Spanish and Japanese). These FAQ archives are collected from customer service or online Q\&A webpage of three companies. The number of question-answer pairs for each archive is 987 (English), 687 (Spanish) and 2384 (Japanese). Each question may have more than one relevant questions within the archives. We split each archive into train, dev and test sets. The number of questions for each set is 790/99/98, 549/69/69 and 1684/350/350.  To train distributed word representations and calculate IDF scores, we employed the English Gigaword (LDC2011T07), the Spanish Gigaword (LDC2011T12) and an in-house Japanese corpus (about 2 billion tokens). These corpus were preprocessed by our in-house tokenizer, and the word2vec\footnote{https://code.google.com/p/word2vec/} toolkit was used for training the distributed word representations.

\subsection{Characteristics of Our Model}
First, we conducted a group of experiments by incrementally adding features. We used the learning-to-rank algorithm to train models, but didn't use hidden layer for these models. Figure \ref{fig:curves}(a) shows the top-1 accuracies on the dev sets. We found that the ``dispersion" and ``penalty" features are effective for English. The ``five important words" features are helpful for both Spanish and Japanese. The ``reverse" feature is very useful for English and Japanese. The ``sparse" features bring around 5\% improvements for all languages.

Second, we verified the effectiveness of our learning-to-rank algorithm. We built two systems: the first one takes only the dense features, and the second one takes both dense and sparse features. The two systems were trained with both the traditional classification method and our learning-to-rank method. Experimental results on dev sets are given in Figure \ref{fig:curves}(b). We see that the learning-to-rank method works consistently better than the classification method.

Third, we illustrated the influence of the neural network structure by changing the number of hidden neurons. We found the model acquired the best performance when using 300 hidden neurons. Figure \ref{fig:curves}(c) shows the performance on the dev sets of two systems. The first system has no hidden layer, and the second one employs 300 hidden neurons.  We can find that using the hidden layer is really helpful for the final performance.

\subsection{Evaluation on the test set}
In this section, we evaluated our systems on the test sets. We tested three systems: (1) ``Dense" takes the dense features, (2) ``Sparse" takes both dense and sparse features, and (3) ``SparseHidden" adds 300 hidden neurons to the second system. We also designed three baseline systems: (1) ``BagOfWord" calculates question similarity by counting how many query words also occur in the candidate; (2) ``IDF-VSM" represents each question with vector space model (VSM) (each dimension is the IDF score of the corresponding word), and calculates the cosine distance as the question similarity; (3) ``Similarity" only uses our ``similarity" feature. Table \ref{tab:testResult} gives the top-1 accuracies. The ``BagOfWord" and "IDF-VSM" systems only counted the exactly matched words, so they didn't work well.  The ``Similarity" system got some improvements by matching words with distributed word representations. The performance of the baseline systems also showed the difficulty of this task. By adding dense and sparse features extracted from word alignment, our systems significantly outperformed the baseline systems.

\subsection{Evaluation on Answer Sentence Selection}
To compare with other state-of-the-art systems, we further evaluated our system on the answer sentence selection task with the standard TREC data set \cite{voorhees1999trec}. The task is to rank candidate answers for each question, which is very similar to our FAQ-based QA task. We used the same experimental setup as \newcite{wang2007jeopardy}, and evaluated the result with Mean Average Precision (MAP) and Mean Reciprocal Rank (MRR). Table \ref{tab:comparison} shows the performance from our system and the state-of-the-art systems. We observe our system get a significant improvement than the other systems. Therefore, our method is quite effective for this kind of ranking tasks.

\begin{table}[tbp]
  \centering
    \begin{tabular}{lccc}
    \toprule
     & English & Spanish & Japanese \\
    \midrule
    BagOfWord &	31.63	&	36.23	&	55.71 \\
IDF-VSM &	37.76	&	37.68	&	58.29 \\
Similarity	&	41.84	&	40.58	&	67.43 \\
Dense	&	45.92	&	44.90	&	67.14 \\
Sparse	&	51.02	&	50.72	&	69.43 \\
SparseHidden	&	52.04	&	59.42	&	70.29 \\
    \bottomrule
    \end{tabular}
    \caption{Evaluation on the test set}
  \label{tab:testResult}
\end{table}

\begin{table}[tbp]
  \centering
    \begin{tabular}{lcc}
    \toprule
     & MAP & MRR \\
    \midrule
    \newcite{wang2007jeopardy} &	0.603 &	0.685 \\
 \newcite{heilman2010tree} &	0.609 &	0.692 \\
 \newcite{yao2013answer} &	0.631 &	0.748 \\
 \newcite{severyn2013automatic} &	0.678 &	0.736 \\
 \newcite{yih2013question} &	0.709 &	0.770 \\
 \newcite{yu2014deep} &	0.711 &	0.785 \\
    \midrule
Our Method &	0.746 &	0.820 \\
    \bottomrule
    \end{tabular}
    \caption{Evaluation on Answer Sentence Selection}
  \label{tab:comparison}
\end{table}

\section{Conclusion}
In this paper, we propose a question similarity model to extract features from word alignment between two questions. We also come up with a bootstrap-based feature extraction method to extract a small set of effective lexical features. By training the model with our learning-to-rank algorithm, the model works very well for both the FAQ-based QA task and the answer sentence selection task.


\end{document}